\documentclass[12pt]{spieman}  
\usepackage{amsmath,amsfonts,amssymb}
\usepackage{graphicx}
\usepackage{setspace}
\usepackage{tocloft}

\usepackage{color,soul}
\usepackage{amssymb}
\usepackage{algorithm}
\usepackage{algorithmic}
\usepackage{amsmath,bm}
\usepackage[utf8]{inputenc}
\usepackage{breqn}
\usepackage{caption,subcaption}
\usepackage{float}
\usepackage{multirow}
\usepackage{booktabs}

\title{Domain Adaptation for Robust Workload Level Alignment Between Sessions and Subjects using fNIRS}

\author[a,$^\dagger$,*]{Boyang Lyu}
\author[b,$^\dagger$]{Thao Pham}
\author[b]{Giles Blaney}
\author[c]{Zachary Haga}
\author[b]{Angelo Sassaroli}
\author[b]{Sergio Fantini}
\author[a]{Shuchin Aeron}

\affil[a]{Tufts University, Department of Electrical and Computer Engineering, 161 College Avenue, Medford, MA 02155}
\affil[b]{Tufts University, Department of Biomedical Engineering, 4 Colby Street, Medford, MA 02155}
\affil[c]{Tufts University, Department of Computer Science, 161 College Avenue, Medford, MA 02155}

\cftpagenumbersoff{figure}
\cftpagenumbersoff{table} 
\begin{document} 
\maketitle

\begin{abstract}
\par\noindent
\textbf{Significance}: We demonstrated the potential of using domain adaptation on functional Near-Infrared Spectroscopy (fNIRS) data to classify different levels of \textit{n}-back tasks that involve working memory. 
\par\noindent
\textbf{Aim}: Domain shift in fNIRS data is a challenge in the workload level alignment across different experiment sessions and subjects. In order to address this problem, two domain adaptation approaches - Gromov-Wasserstein (G-W) and Fused Gromov-Wasserstein (FG-W) were used. 
\par\noindent
\textbf{Approach}: Specifically, we used labeled data from one session or one subject to classify trials in another session (within the same subject) or another subject. We applied G-W for session-by-session alignment and FG-W for subject-by-subject alignment to fNIRS data acquired during different \textit{n}-back task levels. We compared these approaches with three supervised methods - multi-class Support Vector Machine (SVM), Convolutional Neural Network (CNN), and Recurrent Neural Network (RNN). 
\par\noindent
\textbf{Results}: In a sample of six subjects, G-W resulted in an alignment accuracy of 68 $\pm$ 4 $\%$ (weighted mean $\pm$ standard error) for session-by-session alignment, FG-W resulted in an alignment accuracy of 55 $\pm$ 2 $\%$ for subject-by-subject alignment. In each of these cases, 25 $\%$ accuracy represents chance. Alignment accuracy results from both G-W and FG-W are significantly greater than those from SVM, CNN and RNN. We also showed that removal of motion artifacts from the fNIRS data plays an important role in improving alignment performance.
\par\noindent
\textbf{Conclusions}: Domain adaptation has potential for session-by-session and subject-by-subject alignment of mental workload by using fNIRS data. 
 
\end{abstract}


{\noindent \footnotesize\textbf{*}Boyang Lyu,  
\linkable{Boyang.Lyu@tufts.edu} }

{\noindent \footnotesize\textbf{$^\dagger$} These authors contributed equally.}

\begin{spacing}{2}   

\section{Introduction}
Functional near-infrared spectroscopy (fNIRS) is a noninvasive optical technique for monitoring regional tissue oxygenation based on diffusion and absorption of near-infrared light photons in human tissue. Continuous-wave fNIRS provides measurements of concentration changes in oxy-, deoxy- and total-hemoglobin species (\(\Delta[HbO_2]\), \(\Delta[Hb]\), and \(\Delta[HbT]\), respectively) in tissue with temporal sampling rate of on the order of 10 Hz \cite{Franceschini_2000}. Over the past three decades, fNIRS has been found in several brain imaging applications, including non-invasive imaging of cognitive tasks and brain functional activation \cite{Franceschini_2000, Wolf_2002, Bejm_2019,Cui_2012}, and brain computer interface (BCI) \cite{bosworth_2019}. 

Memory-based workload classification using fNIRS measurements has been demonstrated to be an ideal approach for a realistic adaptive BCI to measure human workload level \cite{Hong_2020}. In this paper we study the problem of classification of fNIRS corresponding to different conditions of an \textit{n}-back task (i.e., subjects are required to continuously remember the last \(\textit{n} \in \{1, 2, 3, ...\}\) of rapidly changing letters or numbers). We performed fNIRS measurements on prefrontal cortex (PFC), which has been found to be a relevant area for memory-related tasks by positron emission tomography (PET) and functional magnetic resonance imaging (fMRI) \cite{Owen_2005, Smith_1997}. Most \textit{n}-back classification studies in literature are based on supervised methods on fNIRS signals in within-session and within-subject basis (i.e., within single trial of data acquisition on a single subject) \cite{Herff_2014, Aghajani_2017, Shin_2018}. While those studies showed promising results, subject- and session-dependent systems are not realistic for an interface system that can adapt to different users with a wide range of physiological conditions. With the aim of use in BCI, workload classifications based on fNIRS data across experiment sessions (session-by-session alignment) and across subjects (subject-by-subject alignment) are necessary. 

There are several challenges that hamper accurate workload classification using fNIRS data. We outline them below and propose methods to mitigate them. 

The first challenge, which is the main focus of this paper, is to deal with \textbf{session-by-session} and \textbf{subject-by-subject} variations in classification of \textit{n}-back tasks. These problems are related to what is referred to as domain adaptation in machine learning \cite{ben2010theory,kouw2019review,kurmi2019looking}. More specifically, data from different sessions or different subjects are referred to as belonging to different domains, and the changes in data distributions across different domains (the session or subject that the data belongs to) are considered as a domain shift \cite{shimodaira2000improving}. Due to this phenomenon, the knowledge we learned from one domain cannot be applied directly to another one. To address this problem, recent advances in the theory and methods of optimal transport (OT) \cite{peyre2019computational} and metric measure space alignment \cite{solomon2016entropic, memoli2011gromov,das2018sample} could be used to align data with known labeled \textit{n}-back condition from one session or one subject to the unlabeled data from different session within the same subject or from different subject. Though OT has been applied for domain adaptation with potential performance \cite{courty2016optimal,yair2019optimal}, it has some limitations when two sets of data used for alignment do not share the same metric space, in which case a meaningful notion of distance between two spaces does not exist. For example, for session-by-session alignment, data from some of the fNIRS  channels are removed from one of the two sessions due to a poor signal-to-noise ratio (SNR). This will cause two sessions' data to be embedded in different dimensions in the two domains. A na\"ive solution is to remove the corresponding channels from the other session to guarantee that the two sessions have the same dimension. However, this has a disadvantage of causing loss of information. In this paper, we proposed that using Gromov-Wasserstein (G-W)\cite{memoli2011gromov,peyre2016gromov} and fused Gromov-Wasserstein (FG-W) barycenter \cite{titouan2019optimal} would alleviate this problem and provide algorithms to align across domains for fNIRS \textit{n}-back task classification.

The second challenge is \textbf{motion artifacts} in fNIRS signals. Motion artifacts in fNIRS are commonly due to the coupling changes of any source or detector from the scalp during the experiment. This causes sudden increases or decreases in measured light intensity and can affect the measured fNIRS signals. From a machine learning perspective, motion artifact detection and correction help remove any misleading correlation from the subject behavior (twitching, head movement, etc.) to what the classification model learns from fNIRS data. For example, a classification model may recognize when a subject presses a button as a requirement during the experiment by detecting spikes in the measured signals due to the subject's head movement, instead of detecting real hemodynamic responses from the brain signals. A number of approaches, inspired by statistical signal processing methods such as adaptive filtering, independent component analysis (ICA), and time-frequency analysis, have been proposed to remove or correct for motion artifacts in fNIRS signals \cite{Siddiquee_2018, Zhang_2009, Scholkmann_2010, Cui_2010, Izzetoglu_2010,Medvedev_2008, Sato_2006}. Most of these techniques either depend on the use of auxiliary reference signals (e.g., accelerometry, etc.) or extra optical channels, or require certain assumptions on the characteristics of motion artifacts and cleaned fNIRS signals. In this paper we used an off-the-shelf method based on sparse optimization for automatic detection and removal of spikes and steps anomalies, namely transient artifact reduction algorithm (TARA) \cite{selesnick2014transient}. We will apply the method TARA in the hope to improve classification accuracy of \textit{n}-back tasks. 

The main contributions of this paper to the classification of different \textit{n}-back task conditions include: (1) applying G-W to align fNIRS data during each \textit{n}-back task condition across different experimental sessions for every single subject (session-by-session alignment); (2) applying FG-W barycenter to align fNIRS data during each \textit{n}-back condition between different subjects (subject-by-subject alignment); (3) demonstrating that alignment accuracy could be improved by applying motion artifact removal with TARA as a pre-processing step on fNIRS data. 

\section{Experiment}

\subsection{Subjects and Experiment Design} 

Six healthy human subjects (one female, five males, age range: 23-54 years) participated in this study.  The Tufts University Institutional Review Board approved the experimental protocol, and the subjects provided written informed consent prior to the experiment.

During the \textit{n}-back task, subjects were instructed to watch a series of rapidly flashing random one-digit numbers (from 0-9) shown on a computer screen placed at \textasciitilde{50} cm in front of the subject. Subjects must continuously remember the last \textit{n} numbers (\textit{n} = 0, 1, 2, and 3) and were asked to press the space bar if the currently displayed number (target) matched the preceding \textit{n}-th number. In the 0-back task, the subject pressed the space bar whenever numeral “0” appeared. With increasing \textit{n}, the task difficulty is expected to increase, as the subjects must remember an increasing number of preceding digits and continuously shift the remembered sequence. The experiment was designed such that the targets appeared with 25-35 \% chance (i.e., 65-75 \% non-targets) in each task (chosen randomly). We measured the task performance by counting the number of missed targets (when the subject did not press the space bar for a target), and the number of wrong reactions (when the subject incorrectly identified a non-target stimulus as a target).

\begin{figure}
   \centering
   \includegraphics[width=1\linewidth]{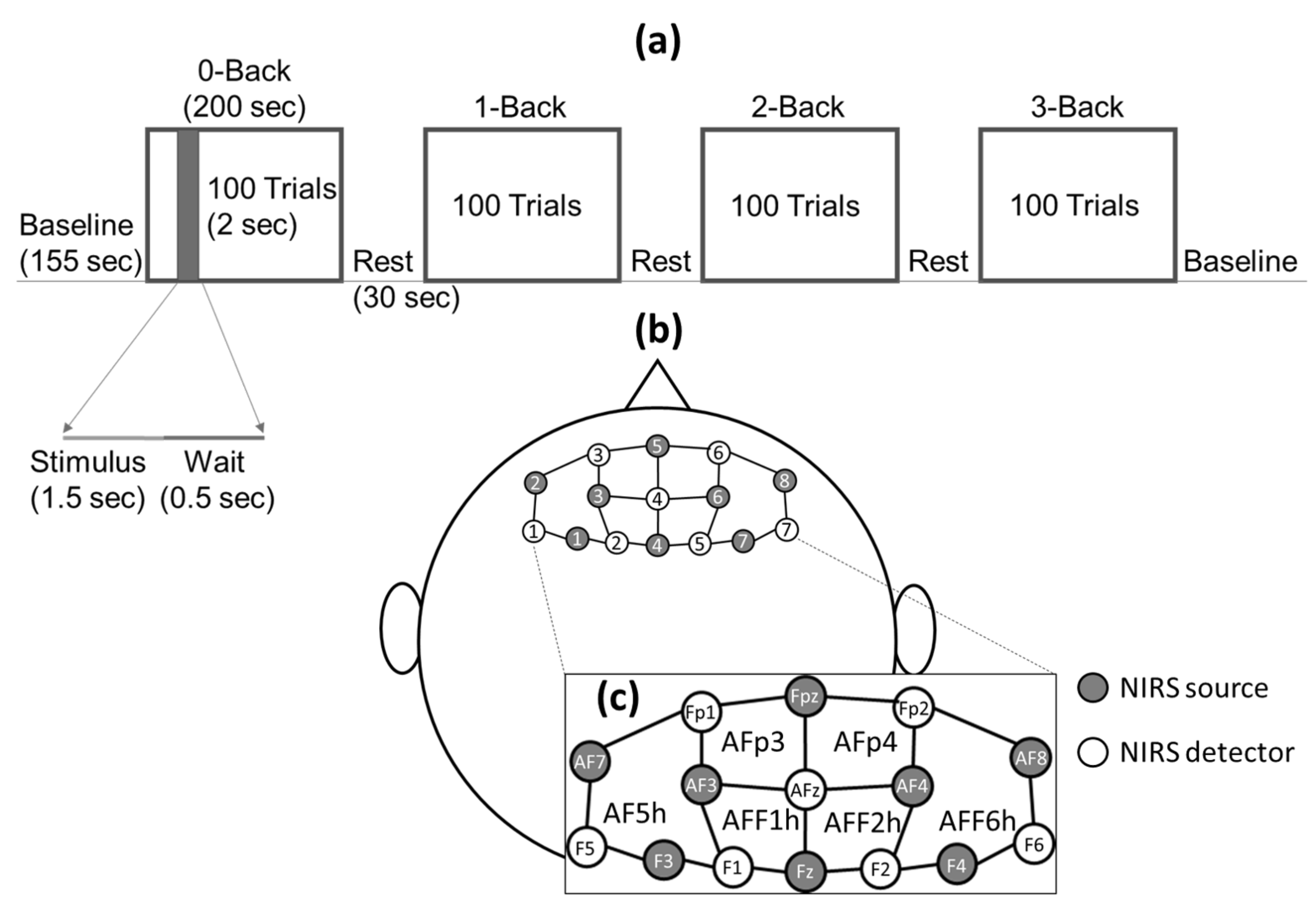}
   \caption{(a) Experimental design for \textit{n}-back task. (b) fNIRS headset with eight sources and seven detectors to give a total of 20 channels at source-detector distance of 3 cm. (c) A zoomed-in view of the schematic in (b) showing positions of 10-10 system (Fp1, Fpz, Fp2, AF7, AF3, AFz, AF4, AF8, F5, F3, F1, Fz, F2, F4 and F6) and 10-5 system (AFp3, AFp4, AF5h, AFF1h, AFF2h, and AFF6h) covered by the sources and detectors.}
   \label{fig:exp_setup}
 \end{figure}

Each subject performed a total of four separate experiment sessions in two days: two sessions per day, one in the morning shift (9-12 a.m.) and one in the afternoon shift (1-4 p.m.). The order of the $n$-back tasks was randomized among sessions, but the randomization order was kept the same among subjects (i.e. only four random sequences were used and each subject was shown each of the four after all of their sessions). A session started with 155 sec of initial baseline with a countdown timer displayed on the screen. At the beginning of a task, an instruction was shown to inform the subject that the upcoming task was 0-, 1-, 2- or 3-back. A task consisted of 100 displayed digits each lasting 2 sec, during which stimulus was displayed for 1.5 sec and followed by a resting time of 0.5 sec where a black screen was shown. Therefore, each task was a total of 200 sec in length. Subsequently, the subject entered 30 sec of baseline (rest) after finishing the task while the performance accuracy of the preceding task was displayed on the screen. This process was repeated for the four values of \textit{n}. At the end of the fourth task, the subject rested for a 155 sec baseline after which the experiment was completed. Figure \ref{fig:exp_setup}a shows the experiment protocol. The entire experiment had a recording time of 20 min (four 200-sec tasks, two 155-sec baselines, and three 30-sec rests in the middle). 

\subsection{Data Acquisition} 
During the entire experiment session, optical data were collected continuously with a continuous-wave fNIRS device (NIRScout, NIRx Medical Technology, Berlin, Germany). Eight light emitting diode (LED) source pairs (at two wavelengths of 760 and 850 nm) and seven detector fiber bundles connected to photodiode (PD) detectors were arranged on a conformable fabric headset. The fNIRS headset can be quickly fixed to the forehead to enable high quality measurements of the prefrontal cortex (PFC) within the range of several minutes. A total of 20 channels at 3 cm source-detector distances were collected. A schematic diagram of the arrangement is shown in Fig. \ref{fig:exp_setup}b, c. Light intensities were collected at a sampling rate of 7.81 Hz. Linear detrending was applied to the collected changes in light intensity with respect to baseline to remove slow temporal drifts. Then the detrended normalized intensities were converted into \(\Delta[HbO_2]\) and \(\Delta[Hb]\) by using the modified Beer-Lambert law \cite{Sassaroli_2004}. We assumed the wavelength-dependent differential pathlength factors (DPFs), which account for the increase in photon pathlength due to multiple scattering, equal to 9.1 and 8.0 for 760 and 850 nm, respectively \cite{Sergiobook}.
\par
During the experiment, continuous arterial blood pressure (ABP) was collected with a beat-to-beat finger plethysmography system (NIBP100D, BIOPAC Systems, Inc., Goleta, CA). ABP measurements were converted into mean arterial blood pressure (MAP, in units of mmHg) and heart rate (HR, in units of beats per minute, bpm).

\subsection{fNIRS Data Pre-processing by TARA}
Measured fNIRS data was checked manually to remove those noisy channels contaminated by high frequency noise ($>1$ Hz). Examples of removed and retained channels from two subjects are shown in Fig.~\ref{fig:bad_channel}, and the number of remaining channels are reported in Table \ref{table:channel_number} in Appendix. The whole session will be removed if more than 60\% of channels are identified as noisy. To further remove motion artifacts from the retained 
channels, we used TARA algorithm \cite{selesnick2014transient}, in which measured time series data are treated as a linear combination of a low-pass signal, motion artifacts and white noise. The algorithm focuses on two types of motion artifacts - transient pulses (spike-like signals) and step discontinuities, and assumes both of them appear infrequently. A sparse optimization problem is then formulated to jointly estimate two types of motion artifacts. We refer the reader to \cite{selesnick2014transient} for more details. We used the code provided by the authors, \footnote{\url{http://eeweb.poly.edu/iselesni/TARA/index.html}} and chose parameters for our fNIRS data as shown in Table \ref{table:TARA_para} in Appendix. Once the motion artifacts are detected, they can be removed from the original signal to obtain the cleaned data.

\begin{figure}[H]
    \centering
    \includegraphics[width=0.8\linewidth]{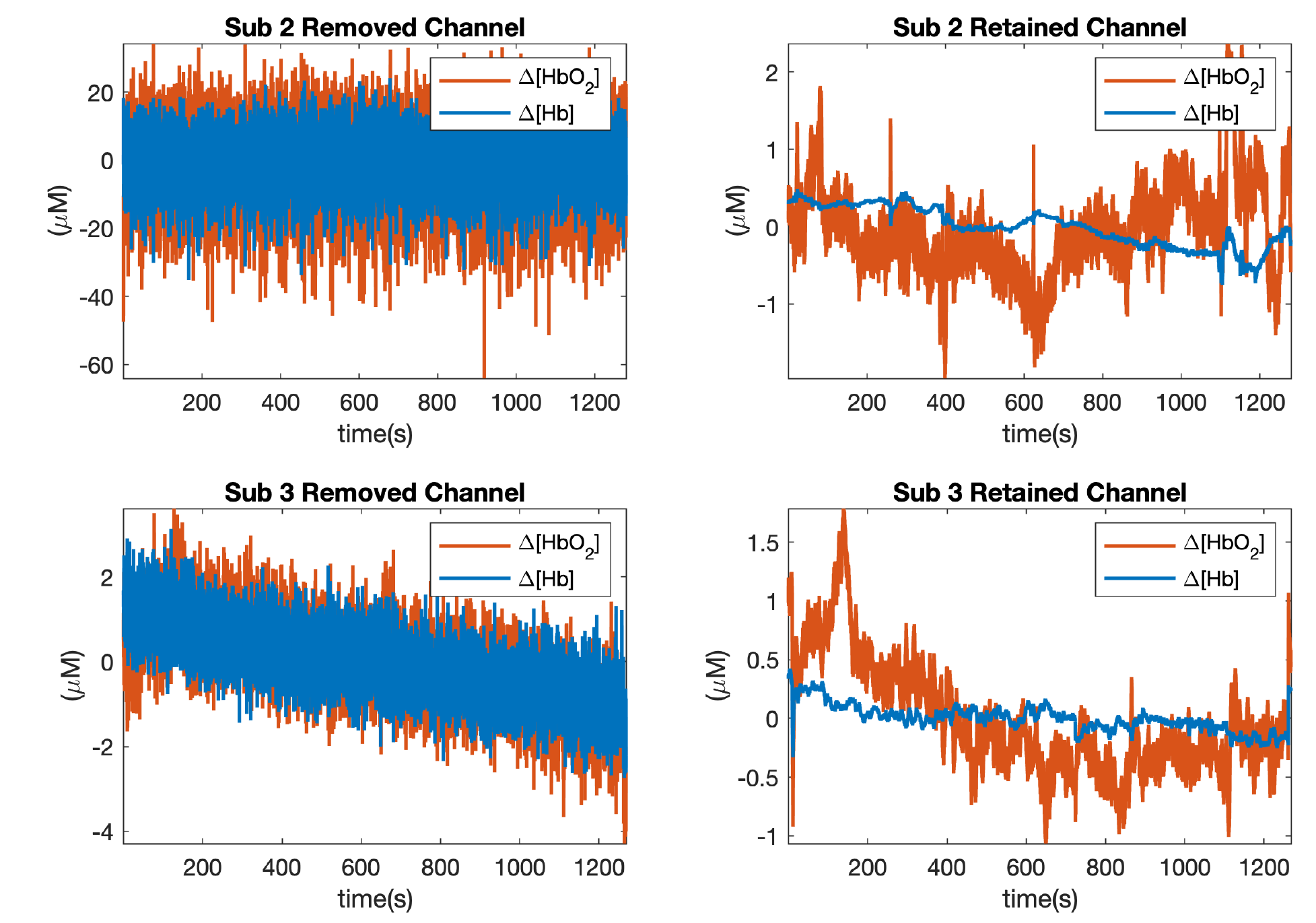}
    \caption{Examples of removed and retained channels from two subjects (2 and 3). The first column shows the removed channels, the second column shows the retained channels. Time courses are shown for concentration changes in oxy-($\Delta[HbO_2]$, shown in \textit{orange}) and deoxy-hemoglobin ($\Delta[Hb]$, shown in \textit{blue}).}
    \label{fig:bad_channel}
\end{figure}

\section{Domain adaptation for fNIRS} 
After the removal of the channels with poor SNR and motion artifacts, a small time duration $w$ is chosen as the window size to divide the remaining \textit{n}-back data (\(\Delta[HbO_2]\) and \(\Delta[Hb]\)) into $N$ non-overlapping small segments. Here we use $w = 60$ samples (\textasciitilde{8}  sec). In order to concretely describe the proposed method, next we will set some notations that are used throughout the paper. 

\paragraph{Notation:} We will use lower-case boldface letters $\bm{x}$ to denote vectors and upper case bold-face letters $\bm{X}$ to denote matrices. Unless otherwise stated, un-bolded lower case letters denote scalars. $\{(\bm{X}_{m,i}^s,y_{m,i}^s)\}_{i=1}^N$ stands for the collection of segmented data set of subject $s$ in its $m_{th}$ session, where $N$ is the number of segments, integer $s\in[1,6]$, and integer $m\in [1,4] $. The $i$-th segment is denoted as $\bm{X}_{m,i}^s\in \mathbb R^{d \times w}$, where $d$ is the number of channels and $w$ is the window length. $y_{m,i}^s \in [0,3]$ is the corresponding \textit{n}-back task label for subject $s$ in session $m$ and segment $i$, $\bm y_m^s = \textbf{vec}(y_{m,i}^s)$ is a $N$ dimensional vector of the label. The remaining notation will be introduced as needed.

\subsection{Session-by-session Alignment} 

\subsubsection{Optimal Transport Theory and Gromov-Wasserstein Matching}
Consider two discrete sets of points $\{\bm{x}_i\}_{i\in 1\cdots n}, \bm x_i \in \mathbb{R}^{d}$ in a metric space $\mathcal{X}$ with a metric $d_{\mathcal{X}}$, and $\{\bm{y}_j\}_{j\in 1 \cdots m},\bm y_j \in \mathbb{R}^{d}$ in another metric space $\mathcal{Y}$ with the metric $d_{\mathcal{Y}}$. The main idea behind aligning two sets of points is by viewing them as two empirical distributions,
\begin{equation}\label{eq:discrete}
\bm{a} = \sum_{i = 1}^{n}  a_i\delta_{\bm{x}_i},  \bm b = \sum_{j = 1}^{m} b_j\delta_{\bm{y}_j}
\end{equation}
where $\delta_{\bm{x}_i}$ and $\delta_{\bm{y}_j}$ are Dirac functions at the position of $\bm{x}_i$ and $\bm{y}_j$, $a_i$ and $b_j$ are the corresponding probabilities. Without further information, $a_i$ and $b_j $ will be set as $\frac{1}{n}$ and $ \frac{1}{m}$ respectively. The Optimal transport (OT) problem is proposed to find a plan $\bm{T} \in \mathbb{R}^{n \times m}$ that is the solution to  
\begin{equation}\label{eq:simple_ot}
\arg \min_{\bm T\in U(\bm a,\bm b)}\langle \bm C,\bm  T\rangle 
\end{equation}
where $\langle \bm C,\bm  T\rangle = \sum_{i,j} \bm C_{i,j} \bm T_{i,j}$, $U(\bm a,\bm b) = \{\bm T \in \mathbb R_+^{n \times m}: \sum_{j = 1}^{m}\bm T_{i,j} = \bm a, \sum_{i = 1}^{n}\bm T_{i,j} = \bm b\}$, $\bm C \in \mathbb R^{n \times m}$ with the $i,j$-th element $\bm C_{i,j }$ being the cost of associating (moving) the point $\bm x_i$ to the point $\bm y_j$. This is also known as the Kantorovich’s relaxation \cite{kantorovich1942transfer} for the original Monge problem \cite{monge1781memoire}. To reduce the computational cost of solving the linear program Eq. \eqref{eq:simple_ot}, an entropic regularization term is usually added to Eq. \eqref{eq:simple_ot}, leading to:  
\begin{equation}
\min_{\bm T\in U(\bm a,\bm b)}\langle \bm C, \bm T\rangle -\lambda H(\bm T)
\end{equation}
where $H(\bm T) = -\sum_{i,j}\bm T_{i, j}(\log \bm T_{i,j} - 1)$. This entropic OT problem \cite{cuturi2013sinkhorn} can be solved efficiently using the Sinkhorn Algorithm \cite{sinkhorn1974diagonal} or its variations such as the Greenkhorn algorithm \cite{abid2018greedy}, both of which can achieve a near-linear time complexity \cite{altschuler2017near}. This approach has been used in domain adaptation \cite{courty2016optimal,yair2019optimal} for transfer of data in different domains. \par
Though widely used for domain adaptation, classic OT lacks the ability of mapping two different metric spaces. When the points have different dimensions, i.e.  $\bm x_i \in \mathbb{R}^{d_1}$ and $\bm y_j \in \mathbb{R}^{d_2}$, where $d_1 \neq d_2$, a distance between $\bm x_i$ and $\bm y_j$ may not be meaningfully defined. Thus, instead of seeking a distance matrix between elements in different domains, Gromov-Wasserstein (G-W) method compares the dissimilarity between the pair-wise distances in each domain. It poses a weaker assumption that if $\bm x_i$ should be aligned to $\bm y_j$ and $\bm x_{i'}$ should be aligned to $\bm y_{j'}$, then for two distance matrices $\bm C^{\mathcal{X}} \in \mathbb{R}^{n \times n}$ and $\bm C^{\mathcal{Y}}\in \mathbb{R}^{m \times m}$ in space $\mathcal{X}$ and $\mathcal{Y}$, $\bm C^{\mathcal{X}}_{i,i'}$ and $\bm C^{\mathcal{Y}}_{j,j'}$ should be similar \cite{solomon2016entropic}. Formally, the G-W distance is defined as 
\begin{equation}\label{gw}
GW((\bm a,\bm C^{\mathcal{X}}), (\bm b,\bm C^{\mathcal{Y}})) = \min_{\bm T\in U(\bm a,\bm b)} \sum_{i,i',j,j'} L( \bm C^{\mathcal{X}}_{i,i'} ,\bm C^{\mathcal{Y}}_{j,j'}) \bm T_{i,j} \bm T_{i',j'}
\end{equation}
where $L$ is a cost function, which typically can be chosen as a quadratic function or Kullback-Leibler divergence. For our method, a squared loss function is applied. Eq. \eqref{gw} is a non-convex problem related to Quadratic Assignment Problem (QAP) \cite{memoli2011gromov}. A regularized version of Gromov-Wasserstein problem is proposed in \cite{solomon2016entropic}, written as
\begin{equation} \label{eq:entropic_gw}
GW_\lambda((\bm a,\bm C^{\mathcal{X}}), (\bm b,\bm C^{\mathcal{Y}})) =\min_{\bm T\in U(\bm a,\bm b)} \sum_{i,i',j,j'} L( \bm C^{\mathcal{X}}_{i,i'} ,\bm C^{\mathcal{Y}}_{j,j'}) \bm T_{i,j} \bm T_{i',j'} - \lambda H(\bm T)
\end{equation}
Problem in Eq. \eqref{eq:entropic_gw} can be solved by projected gradient descent algorithm wherein each iteration solution is found by running Sinkhorn Algorithm \cite{peyre2016gromov}.

\subsubsection{Metric for G-W Alignment for fNIRS data}
For EEG and fNIRS processing, mean and covariance of the time segments have been considered as useful features \cite{barachant2013classification, heger2013continuous}, here we use these features to compute the inner metric matrix of each session. Specifically, for data $\{\bm{X}_{m,i}^s\}_{i=1}^N$ from $m_{th}$ session of subject $s$, we compute its covariance matrices $\{\bm P^s_{m,i}\}_{i=1}^N$ and mean vectors $\{\bm{h}^s_{m,i}\}_{i=1}^N$, where $\bm P^s_{m,i}\in \mathbb{R}^{d \times d}$, $\bm{h}^s_{m,i} \in \mathbb{R}^d$. The distance matrix $\bm C_m^s \in \mathbb{R}^{N \times N}$ is then defined with the $i,i'$-th element $(\bm C_m^s)_{ii'}$ set as

\begin{equation} \label{eq:inner_distance}
(\bm C_m^s)_{ii'} = (\rho_{hellinger}(\bm P^s_{m,i}, \bm P^s_{m,i'}) + \| \bm h^s_{m,i} -  \bm h^s_{m,i'}\|_2) / d
\end{equation}
where $\rho_{hellinger}(\cdot)$ is the matrix version of Hellinger distance \cite{bhatia2019matrix}, written as 

\begin{equation} 
\rho_{hellinger}(\bm A,\bm B) = (tr(\bm A + \bm B) - 2tr(\bm A^{1/2}(\bm A^{-1/2}\bm B\bm A^{-1/2})^{1/2}\bm A^{1/2}))^{1/2}
\end{equation}
where $\bm A$ and $\bm B$ are Positive Definite (PD) matrices. Since the number of channels $d$ selected for different sessions' data are not necessarily the same, we normalize by the number of channels in each session. 

\subsubsection{Domain Adaptation for Session-by-session Alignment}
We assume the label is given for one session's data and aim to infer the label for all other sessions belonging to the same subject. Using the metric defined in Eq. \eqref{eq:inner_distance}, we show the pseudocode for the session-by-session alignment in Algorithm \ref{alg:sess}. Since we only consider data within the same subject, upper index for subject will be dropped in the algorithm.

\begin{algorithm}[H]
\caption{ Alignment between session $m$ and session $n$}
\label{alg:sess} 
\hspace*{\algorithmicindent} \textbf{Input: Source data and label $\{(\bm{X}_{m,i},y_{m,i})\}_{i=1}^N$, target data $\{\bm{X}_{n,i}\}_{i=1}^N$ } \\
\hspace*{\algorithmicindent} \textbf{Output: Target label $\{{y_{n,i}}\}_{i=1}^N$} 
\begin{algorithmic}[1]
\STATE Calculate inner distance matrices $\bm C_m$ and $\bm C_n$ using Eq. \eqref{eq:inner_distance} for $\{\bm{X}_{m,i}\}_{i=1}^N$ and $\{\bm{X}_{n,i}\}_{i=1}^N$ .
\STATE Solve Eq. \eqref{eq:entropic_gw} to get the transport plan $\bm T$ between session $m$ and session $n$.
\STATE Choose the largest value of each column of $\bm T$ as 1 and set others to be 0 to get the coupling matrix $\bm T_{cp}$
\STATE Get target label $\{{y_{n,i}}\}_{i=1}^N$ by calculating $\bm T_{cp}^\top \textbf{vec}(y_{m,i})$
\end{algorithmic}
\end{algorithm} 

\subsection{Subject-by-subject Alignment} 
When targeting subject-by-subject alignment, we assume data and the corresponding labels for all sessions of one subject are given and denote this subject as the source subject. Then we will use these information to predict the labels of fNIRS data for all four sessions of other subjects (target subjects). Transferring labels between different subjects is a bigger challenge since there is a larger shift in domain. Directly using the same G-W alignment as discussed above will lead to a large variance in alignment accuracy. More importantly, we will lose the advantage of knowing all the features and structural information from multiple sessions of the source subject. To address this problem, we consider a recently proposed method named Fused Gromov-Wasserstein (FG-W) \cite{titouan2019optimal}. By computing a FG-W barycenter, which is the Fr\'echet mean of FG-W distance, we summarize all the given information 
into a new representation of the source subject and then follow the same routine as session-by-session alignment to achieve the label alignment.

\subsubsection{Fused Gromov-Wasserstein Barycenter}
Fused Gromov-Wasserstein, unlike the G-W, combines both feature and structural information and shows its advantage in graph classification \cite{vayer2018fused,titouan2019optimal}. Consider two sets of tuples $\{(\bm{x}_i, \bm f_i)\}_{i\in 1\cdots n}$ in space $(\mathcal{X}, \Sigma)$ and $\{(\bm{y}_j, \bm g_j)\}_{j\in 1\cdots m},$ in space $(\mathcal{Y}, \Sigma)$, here $\bm{x}_i\in \mathbb{R}^{d_1}$ and $\bm{y}_j\in \mathbb{R}^{d_2}$ are the data points, $\bm f_i$ and $\bm g_j$ are their corresponding features which are both in space $\Sigma$ and share the same dimension. With a slight abuse of notation, we will use the same symbol as Eq. \eqref{eq:discrete} to denote their empirical distribution, 
\begin{equation}
\bm{a} = \sum_{i = 1}^{n} a_i\delta_{(\bm{x}_i, \bm f_i)},  \bm b = \sum_{j = 1}^{m} b_j\delta_{(\bm{y}_j, \bm g_j)}
\end{equation}
The FG-W distance between such two distributions with both data and the corresponding feature information included is then defined as 
\begin{equation}\label{fgw}
FGW(\bm a, \bm b) = \min_{\bm T\in U(\bm a,\bm b)}  \sum_{i,i',j,j'}  ((1 - \alpha)\rho (\bm f_i, \bm g_j)^q + \alpha | \bm C^{\mathcal{X}}_{i,i'} -\bm C^{\mathcal{Y}}_{j,j'}|^q) \bm T_{i,j} \bm T_{i',j'}
\end{equation}
where $\alpha \in [0,1]$ is a trade-off parameter, $q \geq 1$, $\rho(\bm f_i, \bm g_j)$ stands for the cost of matching feature $\bm f_i$ to feature $\bm g_j$ which in our case corresponds to the labels, i.e. scalar value \textit{n} in the \textit{n}-back task.

For multiple distribution setting like those related to multiple sessions, a natural extension of FG-W distance is its barycenter, which inherits the advantages of FG-W that leverages both structural and feature information. The FG-W barycenter can be obtained by minimizing the weighted sum of a set of FG-W distances. Let $\{\bm C^k\}_{k=1}^K$ be a set of distance matrices, where $\bm C^k \in \mathbb{R}^{N \times N}$, $\{\bm f^k\}_{k=1}^K, \bm f^k \in \mathbb{R^N}$ is the corresponding feature vector. Here $K$ will correspond to the number of sessions for each subject in our case. We assume the base histograms $\{\bm a^k\}_{k=1}^K$ and the histogram $\bm a$ associated with the barycenter is known and fixed as uniform distributions. By calculating the Fr\'echet mean of the FG-W distance, we aim to find a feature vector $\bm f$ and a distance matrix $\bm C$ that represents the structure information, such that
\begin{dmath}\label{eq:fgw_bary}
\min_{\bm C \in{\mathbb{R}^{N\times N}},\bm f \in \mathbb{R}^N,(\bm T^k)_k\in U(\bm a,\bm{a}^k) } \sum_{k}\sum_{i,i',j,j'} \zeta_k ((1 - \alpha) \rho(\bm f_i, \bm f^k_j)^q + \alpha | \bm C_{i,i'} - \bm C^k_{j,j'}|^q) \bm T^k_{i,j} \bm T^k_{i',j'}
\end{dmath}
where $\sum_{k} \zeta_k = 1$ are the weight for sessions and chosen evenly for each session. $q \geq 1$ for the loss of features and a squared loss between features are used for our method. This problem can be solved by Block Coordinate Descent (BCD) algorithm described in \cite{titouan2019optimal}. Note that after solving Eq. \ref{eq:fgw_bary}, only the distance matrix $\bm C$ and feature vector $\bm f$ will be used to form the new representation of the provided $K$ sessions. 

\subsubsection{Metric for FG-W Barycenter Alignment}
Unlike the metric defined in Eq. \eqref{eq:inner_distance} for session-by-session alignment, we removed the L2 norm of the mean difference from the distance when considering the metric for subject-by-subject alignment. This is because the differences of the mean values are usually the same within the same subject but vary across different subjects. It's worth to mention that after removing the L2 norm of the mean difference, the covariance matrices themselves can be viewed as points in a Riemannian space \cite{barachant2011multiclass}. Formally, for the $m_{th}$ session of subject $s$, the distance matrix $\bm C_m^s$ is defined using its covariance matrices $\{\bm P^s_{m,i}\}_{i=1}^N, \bm P^s_{m,i} \in \mathbb{R}^{d\times d}$,  with the $i,i'$-th element  $(\bm C_m^s)_{ii'}$ computed via, 

\begin{equation} \label{eq: hellinger_bary}
(\bm C_m^s)_{ii'} = (\rho_{hellinger}(\bm P^s_{m,i}, \bm P^s_{m,i'})) / d
\end{equation}
\subsubsection{Domain Adaptation for Subject-by-subject Alignment}
The algorithm for subject-by-subject alignment is shown in Algorithm \ref{alg:subject}, here we only take two subjects (each with 4 sessions) as an example but the algorithm can be easily adapted to all other subjects with different number of sessions.
 
\begin{algorithm}[H]
\caption{ Alignment between subject $s$ and subject $t$}
\label{alg:subject}
\hspace*{\algorithmicindent} \textbf{Input: Source data and label $\{(\bm{X}_{\{1...4\}, i}^s,y_{\{1...4\},i}^s)\}_{i=1}^N$, target data $\{\bm{X}_{\{1...4\},i}^t\}_{i=1}^N$ }\\
\hspace*{\algorithmicindent} \textbf{Output: Target label $\{y_{\{1...4\},i}^t\}_{i=1}^N$}
\begin{algorithmic}[1]
\STATE For source and target data, calculate two lists of distance matrices $[\bm C_1^s, \bm C_2^s,\bm C_3^s, \bm C_4^s]$ and $[\bm C_1^t, \bm C_2^t,\bm C_3^t, \bm C_4^t]$ respectively using Eq. \eqref{eq: hellinger_bary}.
\STATE Solve Eq. \eqref{eq:fgw_bary} using $[(\bm C_1^s, \bm y_1^s), (\bm C_2^s, \bm y_2^s),(\bm C_3^s, \bm y_3^s), (\bm C_4^s, \bm y_4^s)]$ to get the inner distance matrix and corresponding label vector of the barycenter for subject $s$, denoted as $\{\bm C^s_{bary}, \bm y^s_{bary}\}$.
\STATE Repeat Step 2 to 4 in Algorithm \ref{alg:sess} using $\{\bm C^s_{bary}, \bm y^s_{bary}\}$ and $\bm C_1^t, \bm C_2^t,\bm C_3^t, \bm C_4^t$ respectively as input to get the labels $\{y_{\{1...4\},i}^t\}_{i=1}^N$ for target data.
\end{algorithmic}
\end{algorithm}

\subsection{Comparison with supervised machine learning methods}


To further demonstrate the potential of domain adaptation methods, we compared our method with a Convolutional Neural Network (CNN), a Recurrent Neural Network (RNN) and a multi-class Support Vector Machine (SVM) based classifiers, applied without any domain adaptation techniques.

For the CNN model, we adapted the architecture structure of EEG-NET \cite{lawhern2018eegnet}. Due to paucity of the amount of data in our case compared to the original paper \cite{lawhern2018eegnet}, we simplified the structure to three convolutional layers followed by two dense layers. Details of the CNN structure can be found in Appendix, Table \ref{cnn_strucuture}. To compare with session-by-session alignment using the G-W method, \(\Delta[HbO_2]\) and \(\Delta[Hb]\) data were first separated and then stacked along a new dimension as input to the CNN. Since the removal of some noisy channels will lead to different input data points being in different dimensions, thereby causing a mismatch between input data and the fixed model structure, we replaced the discarded channels with the average of data from the remaining channels (separately for \(\Delta[HbO_2]\) and \(\Delta[Hb]\)). We used data from one session as input to train the model with Adam optimizer \cite{kingma2014adam} using cross entropy loss and tested on the remaining sessions. The model was trained until severe overfitting happened (300 epochs in our case) to guarantee the best test accuracy can be achieved within the training process. Test accuracy was recorded during the whole training process (i.e. after each training epoch) and the best result was selected among them. The training and testing processes were conducted 5 times and the average test accuracy was reported. To compare with subject-by-subject alignment using FG-W method, data from all four sessions of one subject were combined and used as input and the classification model was trained to predict the task labels for all other five subjects in the same manner as discussed above.
\par
For the RNN model, we used a basic three-layer Long Short Term Memory (LSTM) \cite{hochreiter1997long} model with the hidden size set as 20. The training and testing data were prepared in the same way as the CNN method except that \(\Delta[HbO_2]\) and \(\Delta[Hb]\) data were not separated but input together. For both session and subject prediction, the training and evaluation procedure followed the same routine as CNN.
\par
Before applying SVM, a dimension reduction technique was applied to the segmented multi-channel fNIRS data. Here we used UMAP (Uniform Manifold Approximation and Projection) \cite{mcinnes2018umap} to compress each piece of segmented data  \cite{ali2019timecluster} into a 50 dimensional vector, with the distance matrix calculated using Eq. \ref{eq:inner_distance}
for session-by-session alignment and Eq. \ref{eq: hellinger_bary} for subject-by-subject alignment. During the training procedure for session-by-session alignment, only one session's data was used for testing and all the remaining data was used for training. Hyper-parameters were selected by leave-one-session-out cross-validation. This was similar for subject-by-subject alignment, one subject's data (including all the sessions) was used for testing while all other data was used for training. Hyper-parameters were again selected by leave-one-subject-out cross-validation.

\par
The Student's t-test was used to investigate differences between alignment accuracy from G-W/FG-W methods with 25$\%$ chance levels ($25 \%$ stands for the chance to assign any session as 0, 1, 2 or 3-back), between G-W/FG-W methods using raw and cleaned data from TARA, and between G-W/FG-W and comparison methods stated above. All values are reported as mean $\pm$ standard error weighted by the standard deviations of the alignment accuracy values from six subjects unless otherwise noted.

\section{Results}
\subsection{Subject Performance} 
\label{subsec:sub_performace}
\begin{figure}[h]
   \centering
   \includegraphics[width=1\linewidth]{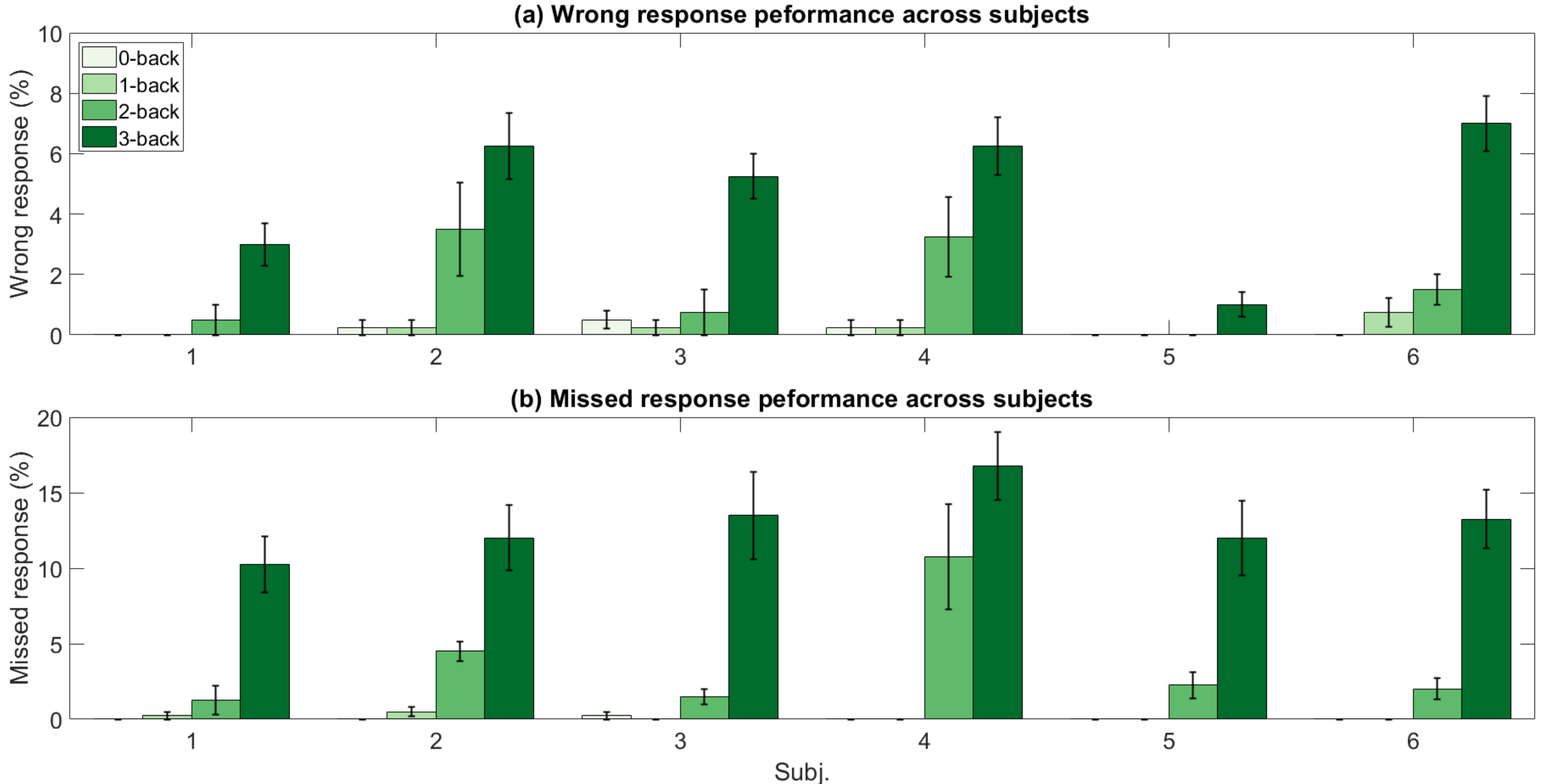}
   \caption{Summary of subject performance for the \textit{n}-back task: average percentages of wrong responses (a) and missed responses (b) for \textit{n}-back task conditions across subjects. Bars represent the means, and error bars represent standard errors across four experimental sessions.}
   \label{fig:sub_performance}
\end{figure}

 Figure \ref{fig:sub_performance} shows summary of subject performance analysis with the average percentages of wrong and missed responses, respectively, across four sessions and six subjects for each \textit{n}-back task condition. The difficulty level, in terms of the amounts of wrong and missed responses, increases significantly for the 3-back task as compared to other \textit{n}-back tasks ($p<0.05$, paired-sample $t$-test). Next, the amounts of wrong and missed responses for 2-back task in the four experiment sessions are significantly higher than those for the 1- and the 0-back task ($p<0.05$, paired-sample $t$-test). Finally, there was no significant difference in the difficulty level between the 0- and the 1-back task in terms of wrong and missed responses ($p>0.05$, paired-sample $t$-test).          

\subsection{Peripheral physiological measurements}
\label{subsec:physiol_eval}
\begin{figure}[H]
   \centering
   \includegraphics[width=1\linewidth]{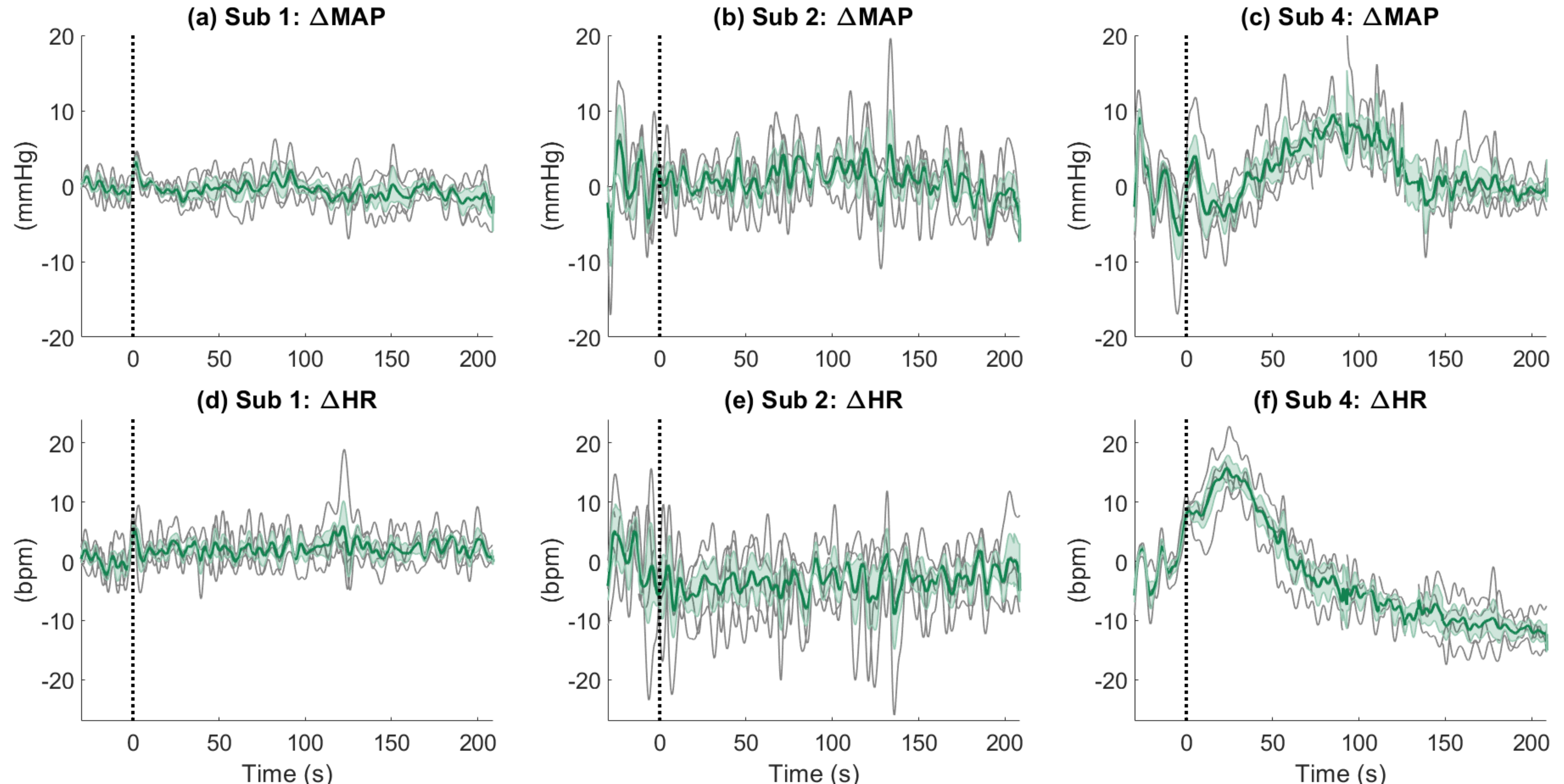}
   \caption{Average changes in hear rate (HR) and mean arterial pressure (MAP) across all sections of 2-back task for three subjects (1, 2 and 4). The time traces are shown starting from 30 sec before the task. Black dotted lines indicate time $t=0$ s. Solid green lines are the averaged across sessions in $\Delta\text{MAP}$ and $\Delta\text{HR}$; standard errors of these averages are shown by the cyan shaded regions; solid grey lines depict the individual measurements.}
   \label{fig:physiol}
\end{figure}

Figure~\ref{fig:physiol} presents the examples of average time courses of changes in MAP and HR from three subjects (1, 2 and 4) across different measurement sessions of 2-back task. We observe a greater variability in task-evoked changes in HR and MAP across subjects than across sessions. In particular, subjects 1 and 2 show negligible changes in MAP and HR during the task with respect to the initial baseline, with individual measurements from different sessions following the same trend. On the other hand, all the measurements from different sessions from subject 4 show totally different responses as compared to subjects 1 and 2. For subject 4, MAP increases during the middle of the task and then returns to the baseline in the last minute. The HR measurements from this subject feature an initial increase and immediate decrease at the onset of the task.

\subsection{Effects of Motion Artifact Removal using TARA} 

\begin{figure}
    \centering
    \includegraphics[width=0.8\linewidth]{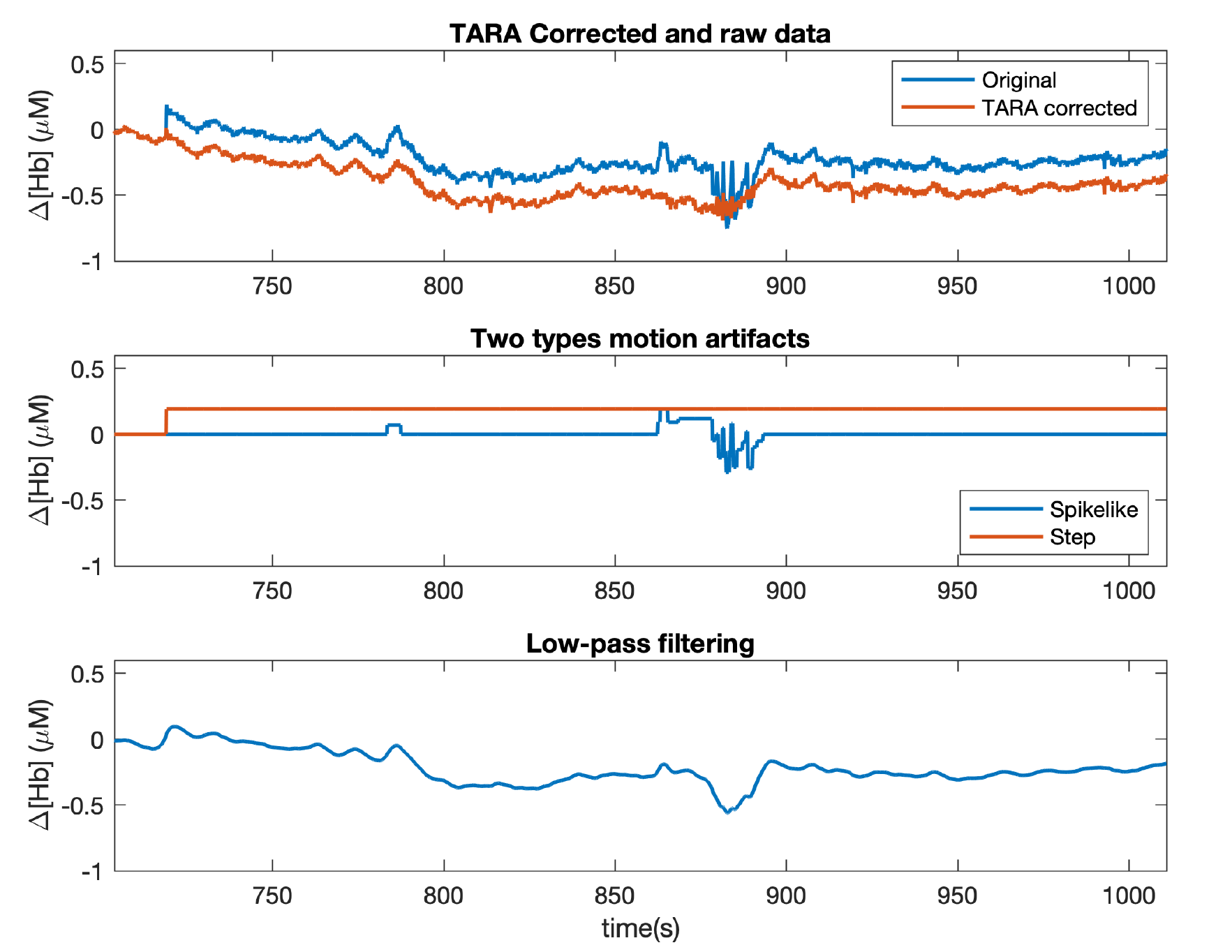}
    \caption{Effect of using TARA on \(\Delta[Hb]\) signal. The top row shows the original and cleaned signals. The second row shows the detected motion artifacts, including spike- and step-like features. The cleaned signal is obtained by subtracting these motion artifact features from the original signal. The last row shows the low-pass filtered signal of the original data. Distortion in the signal arisen from the step discontinuity can still be observed from the low-pass filtered signal.}
    \label{fig:motion_artifact_remov}
\end{figure}

Figure \ref{fig:motion_artifact_remov} displays the effects of TARA in removing motion artifacts in fNIRS signals (\(\Delta[Hb]\)). As shown in the figure, the original signal is contaminated by the motion artifacts with spikes and steps. After applying TARA algorithm, most of the motion artifacts have been removed. As compared to applying low-pass filter to the original signal, TARA does not bring any further distortion to the cleaned signal and is more effective at removing step artifacts. The effect of this motion artifacts removal as a pre-processing step before applying alignment algorithms is also shown in Table \ref{table:combined_sess_and_sub} 
and Fig. \ref{fig:Table1plot}. An improvement for session-by-session alignment accuracy (by an average of $3 \pm 3\%$ across six subjects; $p<0.005$, paired-sample t-test) and subject-by-subject alignment accuracy (by an average of $5 \pm 2\%$ across six subjects; $p<0.0005$, paired-sample t-test) can be seen after applying TARA on fNIRS signals.

\subsection{Session-by-session Alignment}\label{section: sess_by_sess}
\begin{figure}
    \centering
    \includegraphics[width=0.5\textwidth]{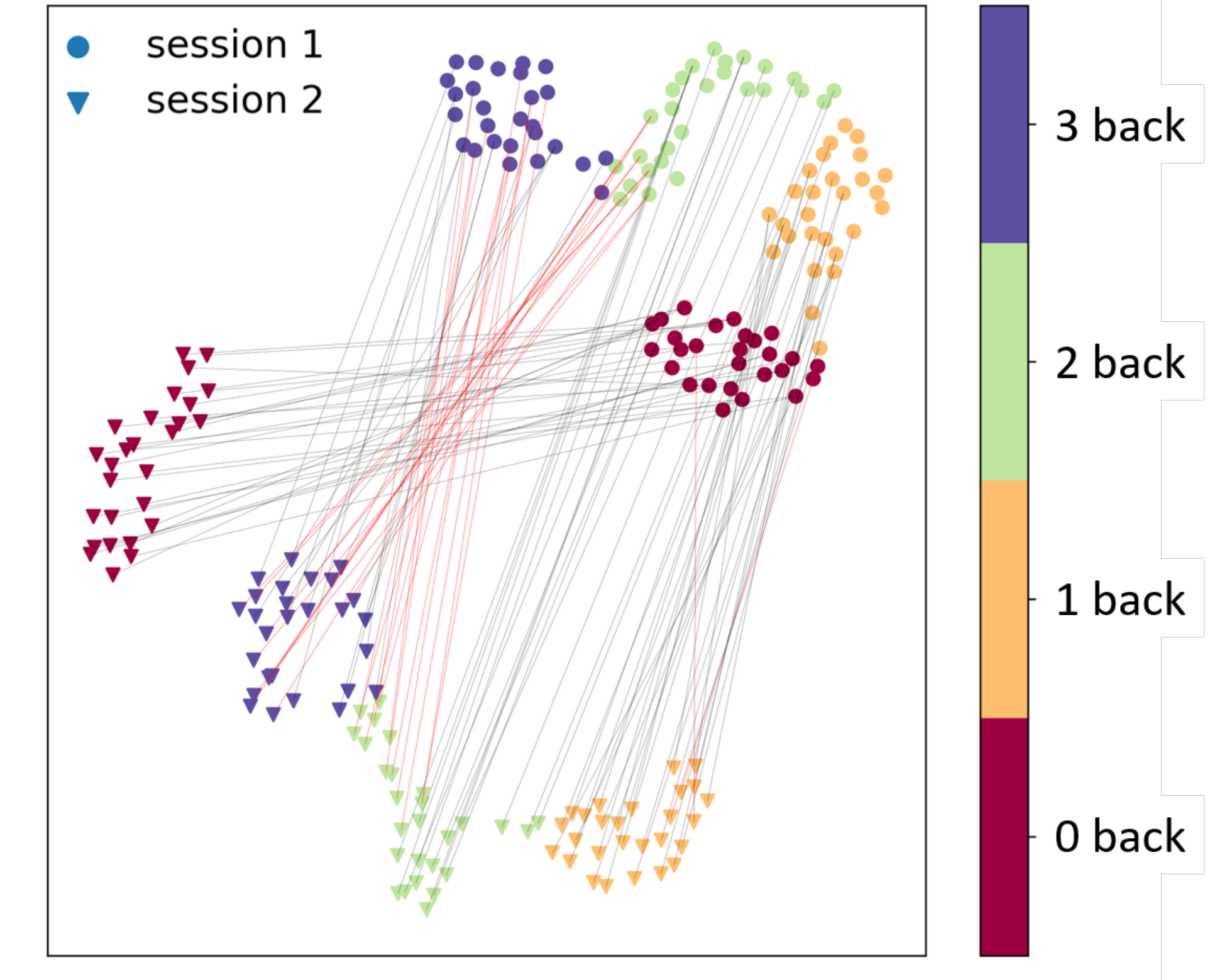}
    \caption{Visualization of the alignment from session 1 to session 2 for subject 4. Circles indicate data from session 1 and triangles indicate the data from session 2. Four different colors represent 0- to 3-back experiments. Black lines indicate correct alignment and red lines indicate misalignment.}
    \label{fig:sess_align_show}
\end{figure}

A low dimensional UMAP visualization of the alignment for two sessions' data is shown in Fig.~\ref{fig:sess_align_show} for subject 4. In Fig.~\ref{fig:sess_align_show}, the low dimensional projection was generated individually from the distance matrix of each session's data. Therefore, the positions of the two groups of sessions' data are assigned randomly and their relative distances are not their true distances.\par

\begin{figure}
 \centering
    \includegraphics[width=1\linewidth]{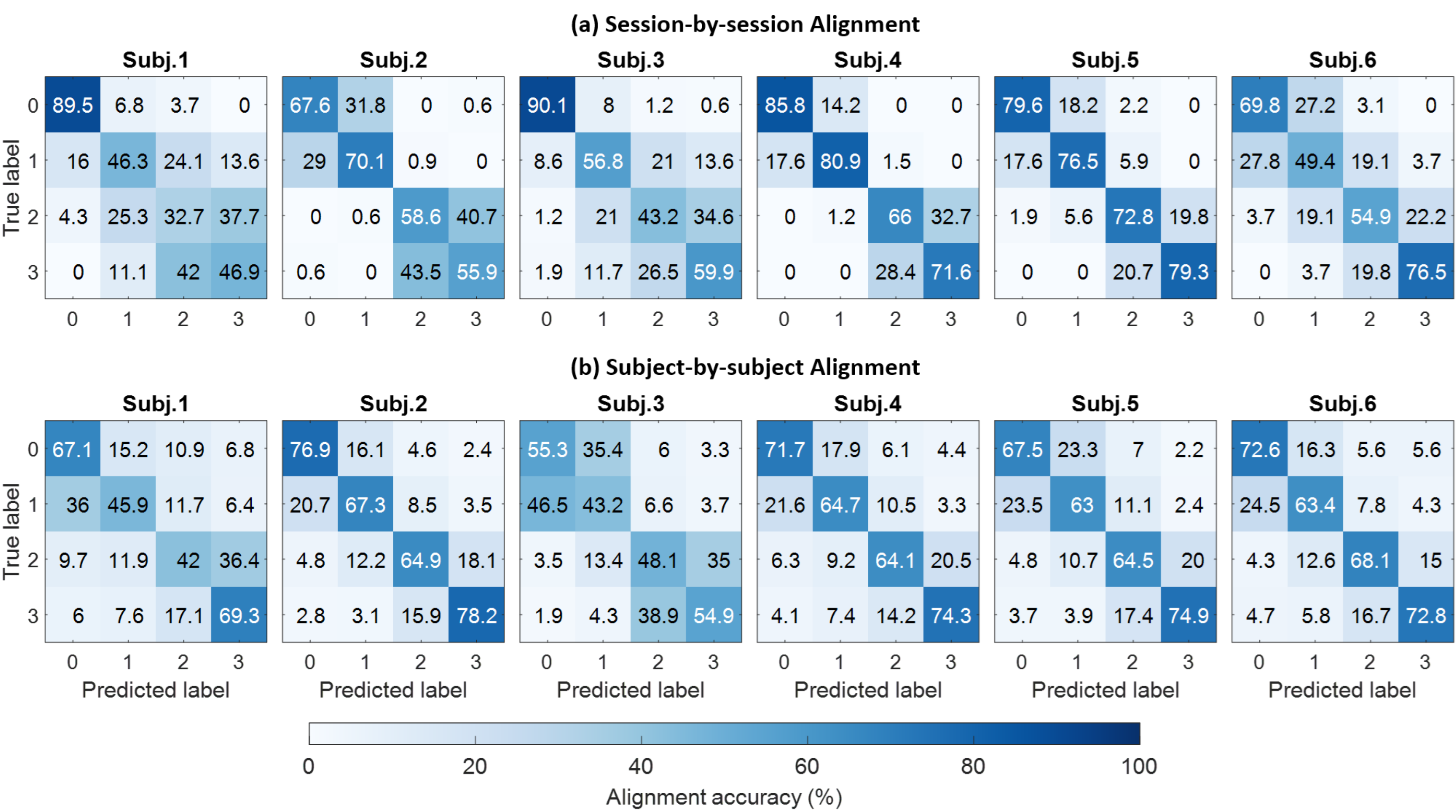}
    \caption{Confusion matrices of session-by-session and subject-by-subject alignments in six subjects. The first row (a) presents session-by-session alignment accuracy within each subject. Each number reported in each confusion matrix is the average accuracy from the alignment of every two separate sessions among four sessions.  The second row (b) presents subject-by-subject alignment accuracy. Each number reported is the average accuracy from the alignment between one source subject to other five target subjects.}
\label{fig:session_and_subject_conf_mat}
\end{figure}

\begin{figure}
 \centering
    \includegraphics[width=1\linewidth]{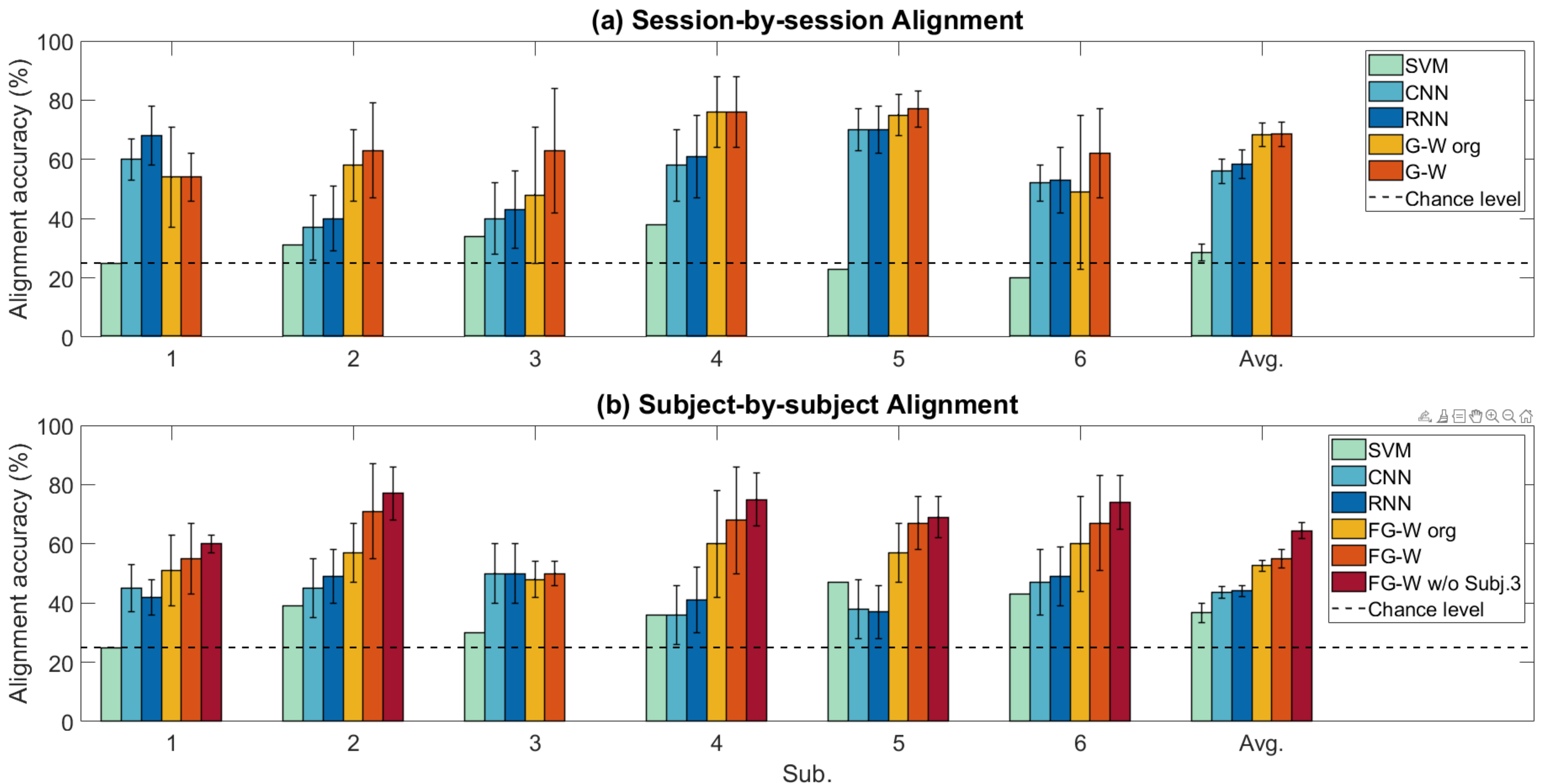}
    \caption{Average alignment accuracy ($\%$) from six subjects. For session-by-session alignment (a), values are shown for SVM, CNN and RNN using cleaned data from TARA, Gromov-Wasserstein (G-W) using original data and cleaned data from TARA. For subject-by-subject alignment (b), values are 
   shown for SVM, CNN and RNN using cleaned data from TARA,  Fused Gromov-Wasserstein (FG-W) using original and cleaned data from TARA, and FG-W using cleaned data from TARA when data from subject 3 is excluded. Bars represent the means, and error bars represent standard errors.}
\label{fig:Table1plot}
\end{figure}

Figure \ref{fig:session_and_subject_conf_mat} presents the confusion matrices of session-by-session alignment for four \textit{n}-back tasks (0, 1, 2, and 3) of six subjects. Numbers reported in the confusion matrix are the average alignment accuracies of all the possible combinations of two out of all four sessions for each subject. Values in the main diagonal of each confusion matrix represent correct alignment between predicted and true label, while the other values represent the misalignment results. Correct alignment results are significantly greater than chance level of 25$\%$ ($p<0.0001$, one-sample $t$-test). The averages and standard deviations of session-by-session alignment accuracy for six subjects are summarized in Table~\ref{table:combined_sess_and_sub} and Fig.~\ref{fig:Table1plot}. For each subject, the value reported is calculated based on the alignment or prediction accuracy for all possible combination of session pairs. 
As compared to SVM, CNN and RNN, alignment accuracy of G-W is greater by an average of $43 \pm 5\%$, $7 \pm 4\%$, and $5\pm5\%$, respectively ($p<0.005$, paired-sample t-test).

\subsection{Subject-by-subject Alignment} 

Figure \ref{fig:session_and_subject_conf_mat} displays the confusion matrices of subject-by-subject alignment for four \textit{n}-back tasks of six subjects. Each number in the reported confusion matrix is the average of alignment accuracies of different tasks from the source subject to five other subjects as the targets. Correct alignment results are significantly greater than chance level of 25$\%$ ($p<0.0001$, one-sample $t$-test). Average subject-by-subject alignment accuracy is shown in Table~\ref{table:combined_sess_and_sub} and Fig.~\ref{fig:Table1plot}. Each reported average accuracy value is the average of the alignment accuracy when considering one subject as the source and five other subjects as the targets. For each subject pair, accuracies are calculated between source subject and all sessions within target subject and averaged to obtain the accuracy between source and target subject. As compared to SVM, CNN and RNN, alignment accuracy of FG-W is greater by an average of $22\pm 
2\%$, $15\pm 5\%$, and $15\pm5\%$, respectively ($p<0.0005$, paired-sample t-test).
\par
Since data from subject 3 has poor SNR and is severely affected by motion artifacts in half of the fNIRS data (see appendix Table~\ref{table:channel_number}), we could treat this subject as an outlier. Alignment accuracy without using data from subject 3 is reported in Table~\ref{table:combined_sess_and_sub} and Fig. \ref{fig:Table1plot}.

\begin{table}
\caption{Average session-by-session and subject-by-subject alignment accuracy ($\%$) by using G-W and FG-W, respectively, as compared with SVM, CNN and RNN. G-W and FG-W barycenter alignment methods were applied to both original data (org) and data cleaned by TARA algorithm, by using data including and excluding data from subject 3. For other methods, only cleaned data were used as input. Averages and standard errors across all subjects are reported (Avg.).}
  \label{table:combined_sess_and_sub}
  \centering
  \begin{center}
  \begin{tabular}{|c|c|c|c|c|c|c|c|c|}
    \hline\hline
        \multicolumn{2}{|c|}{}  & Sub 1 & Sub 2 & Sub 3 & Sub 4 & Sub 5 & Sub 6 & Avg.\\\hline\hline
    \multirow{5}{*}{Sess-by-sess}
     & SVM      &25 & 31 & 34 & 38 & 23 & 20 & 29 $\pm$ 3$^{**}$\\
     \cline{2-9} 
     & CNN      &60$\pm$7 & 37$\pm$11 & 40$\pm$12 & 58$\pm$12 & 70$\pm$7 & 52$\pm$6 & 56 $\pm$ 4$^{**}$\\
     \cline{2-9} 
     &RNN       &68$\pm$10 & 40$\pm$11 & 43$\pm$13 & 61$\pm$14 & 70$\pm$8 & 53$\pm$11 & 58 $\pm$ 5$^{*}$\\
     \cline{2-9} 
     & G-W org  &54$\pm$17 & 58$\pm$12 & 48$\pm$23 & 76$\pm$12 & 75$\pm$7 & 49$\pm$26 & 68 $\pm$ 4$^{*}$\\
     \cline{2-9} 
     & G-W      &54$\pm$8 & 63$\pm$16 & 63$\pm$21 & 76$\pm$12 & 77$\pm$6 & 62$\pm$15 & 68 $\pm$ 4\\
    \hline\hline
    \multirow{6}{*}{Sub-by-sub}
     & SVM      &25 & 39  & 30  &  36  & 47 & 43 & 37 $\pm$ 3$^{**}$\\
     \cline{2-9} 
     & CNN      &45$\pm$8 & 45$\pm$10 & 50$\pm$10 & 36$\pm$10 & 38$\pm$10 & 47$\pm$11 & 44 $\pm$ 5$^{**}$\\
     \cline{2-9} 
     &RNN       &42$\pm$6 & 49$\pm$9 & 50$\pm$10 & 41$\pm$11 & 37$\pm$9 & 49$\pm$10 & 44 $\pm$ 2$^{**}$\\
     \cline{2-9}
     & FG-W org &51$\pm$12 & 57$\pm$10 & 48$\pm$6 & 60$\pm$18 & 57$\pm$10 & 60$\pm$16 & 53 $\pm$ 2$^{**}$\\
     \cline{2-9} 
     & FG-W     &55$\pm$12 & 71$\pm$16 & 50$\pm$4 & 68$\pm$18 & 67$\pm$9 & 67$\pm$16 & 55 $\pm$ 2\\
     \cline{2-9}
     & FG-W w/o &60$\pm$3 & 77$\pm$9 & N/A & 75$\pm$9 & 69$\pm$7 & 74$\pm$9 & 64 $\pm$ 3\\
     & Sub 3& & & & & & &\\
     \hline\hline
  \end{tabular}
  \footnotesize{$^*$ $p$ $<$ $0.005$ compared to G-W or FG-W, $^{**}$ $p$ $<$ $0.0005$ compared to G-W or FG-W}\\
    \end{center}
\end{table} 

\subsection{Combining \textit{n}-back tasks in Session-by-session and Subject-by-subject Alignment}
The analysis of subject performance (Sec.~\ref{subsec:sub_performace}) showed significant differences in the number of missed targets and wrong reactions depending on the \textit{n}-back task conditions. Particularly, subject performance suggests that 0- and 1-back tasks could be combined together in the alignment since they show similar brain activation behaviors. In this section, we showed that by combining data from 0-back together with 1-back, and 2-back together with 3-back tasks, the alignment accuracy increased abruptly for both session-by-session and subject-by-subject alignment, as shown in Table \ref{table:combine01_23}. As compared to results reported in Table \ref{table:combined_sess_and_sub}, session-by-session alignment accuracy increased by an average of $22 \pm 
2\%$ , and subject-by-subject alignment accuracy increased by an average of $33 \pm 3\%$ .  

\begin{table}[h]
\caption{Average session-by-session and subject-by-subject alignment accuracy ($\%$) from G-W and FG-W methods, respectively, when combining 0-back together with 1-back tasks, and 2-back together with 3-back tasks. Cleaned fNIRS data were used. Averages and standard errors across all subjects are reported (Avg.).}
  \label{table:combine01_23}
  \centering
  \begin{center}
  \begin{tabular}{|c|c|c|c|c|c|c|c|}
  \hline\hline
   &Sub 1 & Sub 2 & Sub 3 & Sub 4 & Sub 5 & Sub 6 & Avg.\\
  \hline
   Sess-by-sess using G-W &  $79\pm8$  &  $99\pm2$  &   $82\pm17$  &  $99\pm1$ &  $96\pm2$ & $87\pm 10$  & $98\pm3$\\
   \hline\hline
   Sub-by-sub using FG-W  &  $81\pm11$  &  $89\pm9$  &   $89\pm3$  &  $86\pm13$  &  $88\pm7$ & $86\pm13$ & $88\pm2$\\
   \hline\hline
  \end{tabular}
  \end{center}
\end{table} 

\section{Discussion}
In this study of six subjects, we showed that fNIRS signals measured from 20 channels on the PFC can be used to robustly discriminate subjects' mental workload between different \textit{n}-back task levels across sessions within one subject and across different subjects. One limitation of our study is a small number of subjects. However, with the current number of subjects (six subjects), this paper still achieved the goals of demonstrating: (1) an alignment accuracy greater than that of chance (25\%) for the majority of session-session and subject-subject combinations; and (2) greater accuracies than that obtained from multi-class SVM, CNN and RNN based model. We thereby showed the potential of fNIRS as a modality for BCI and user state monitoring that can adapt to different users with various physiological states. Future works will address the extension of this study with a larger sample of subjects to further investigate the variability between sessions and subjects.
\par
In regards to data pre-processing, we show that motion artifact removal in fNIRS signals is an important step for the following mental workload alignment. Specifically, we report that using TARA to remove motion artifacts from fNIRS signals increased alignment accuracy by an average of $3 \pm 3 \%$ for session-by-session alignment and by $5 \pm 2 \%$ for subject-by-subject alignment ($p<0.005$). Future work could include addressing different types of artifacts that could arise in fNIRS time series which were not considered by TARA, such as oscillatory transients. Additionally, possible future improvements in TARA may be to investigate an automatic way for selection of regularization and non-convexity parameters in TARA algorithm across subjects.
\par

We introduced two approaches, G-W and FG-W barycenter, for session-by-session and subject-by-subject alignment of mental workload during \textit{n}-back task. We proved that our methods could be generalized across different sessions and subjects data. In particular, for session-by-session alignment, we used labeled fNIRS data with known \textit{n}-back task conditions from one session to align with other unlabeled sessions from the same subject by using G-W method. We showed that most of the unlabeled sessions' data could be mapped correctly to their true labels, with the alignment accuracy ranging from $54$ to $77 \%$ (with $25 \%$ representing chance alignment). Meanwhile, with multi-class SVM and simplified CNN and RNN model, the \textit{n}-back task classification accuracy was lower (by an average of $43\pm5\%$, $7 \pm 4 \%$, and $5\pm5\%$, respectively). Note that CNN and RNN required the same amount of data as the proposed methods for training, while SVM required more data (from more than one session) for training. Similar for subject-by-subject alignment, we used labeled data from one subject as the source data for alignment. Labels and structural information of the source data were combined to generate a new representation (i.e. the FG-W barycenter). Following the same routine as the session-by-session alignment, we were able to use the barycenter from the source subject to predict the labels for data from different sessions for other subjects, with the alignment accuracy from $50$ to $71 \%$ (also with $25 \%$ representing chance alignment). From the corresponding SVM, RNN and CNN methods, \textit{n}-back task classification performance achieved lower accuracy than FG-W method (by an average of 15 to 22 $\%$). Again, CNN and RNN were trained from data from one subject (source data), while SVM was trained from data from five subjects for classification. Moreover, our methods of G-W and FG-W do not require the two subsets of data used for alignment to have the same dimension. Thus they do not require data interpolation due to removing noisy fNIRS signals as for CNN and RNN method. However, we note that even though G-W and FG-W methods are free from the dimension requirement for data, they could not achieve satisfying results when a large amount of data is missing (e.g., in the case of subject 3 when around half of the channels were discarded in the pre-processing step).
\par
We found relatively higher alignment results for session-by-session alignment (average of $68 \pm 4 \%$) than subject-by-subject alignment (average of $55 \pm 2 \%$). One source of variation in fNIRS data across experiment sessions and across different subjets could come from the variability in systemic physiology, as seen in the variability in the task-evoked changes in MAP and HR (see Sec.~\ref{subsec:physiol_eval}). We observed that the variability of these two physiological measurements are larger across subjects than across sessions. This may explain the a greater accuracy results for session-by-session alignment than subject-by-subject alignment. Another source of variability across sessions and subjects may also come from the variation in fNIRS optode placement on the subject's head. We anticipate the optode placement variation to be greater across subjects than across sessions due to different head geometry from different human subjects. From our results, we found that the new representation of the barycenter of the source subject still aligned well to data from other subjects even though subject-by-subject alignment was a more challenging problem. This is indicative that representations of different subjects may still share similar underlying structures even from different domains. Future work will explore generating barycenter from source data from multiple subjects' information for subject-by-subject alignment to account for the across-subject variations in the barycenter.
\par
Based on our alignment results shown in confusion matrices in Fig. \ref{fig:session_and_subject_conf_mat}, the misalignment in session-by-session and subject-by-subject alignment are relatively high between 0-back and 1-back, and between 2-back and 3-back tasks. In particular, the misalignment is the highest between 2- and 3-back task (when 2-back task is the true label and 3-back is the predicted label and vice versa), ranging from $19.8$ to $43.5 \%$. The second highest misalignment is between 0- and 1-back task, ranging from $6.8$ to $31.8 \%$. Similarly, for subject-by-subject alignment, the highest misalignment came from 0- and 1-back task, ranging from $15.2$ to $46.5 \%$. The second highest misalignment is between 2- and 3-back, ranging from $14.2$ to $38.9 \%$. This gave us an idea of combining 0- with 1-back tasks, and 2- with 3-back tasks in the alignment. Substantial increases in alignment performance (by an average of $22 \pm 2 \%$ for session-by-session and $33 \pm 3 \%$ for subject-by-subject alignment) suggest that future works could study workload classification between rest to low workload level (0- and 1-back tasks) versus high workload level (2- and 3-back tasks).
\par
Finally, single-distance continuous-wave (CW) fNIRS measurements of intensity from source-detector pairs at 3 cm distance were used in this study. This measurements have been known to be more sensitive to hemodynamic changes in superficial tissues (i.e., scalp and skull) than in the brain \cite{Tachtsidis_2016}. Previous study \cite{kirilina2012physiological} has shown that tasked-evoked superficial artefacts may arise during brain activation task due to systemic changes in peripheral physiology rather than the cerebral hemodynamics. This also confirm the claim that variations in our alignment results across sessions and subjects could be partially due to variability in systemic physiological origins. For the purpose of our aims, it is desirable to increase the sensitivity of our measurements to brain tissue, in order to probe hemodynamic changes associated with brain activation. One approach, namely the dual-slope method, involves a simple implementation of a certain arrangement of sources and detectors to localize sensitivity of NIRS measurements to a deeper region \cite{Sassaroli:s}, thus suppressing confounding signals from superficial tissue. This approach could also help remove instrumental drifts and motion artifacts from measured signals as dual-slopes are unaffected by changes in optical coupling. Future extensions of this work may involve implementing the dual-slope configuration in such experiments as those described here. Another approach to correct for extracerebral contamination is to acquire measurements in a multi-distance arrangement to incorporate short ($<$1 cm, sensitive to extracerebral tissue only) and long ($>$2.5 cm, sensitive to both extracerebral and brain tissues) source-detector separations \cite{FantiniAPL}, and apply processing method such as adaptive filtering \cite{zhang2007adaptive} to remove global interference from systemic physiology from fNIRS measurements.  

\section{Conclusions}

In order to illustrate that fNIRS signals can be effectively used to identify subjects’ mental workload between different $n$-back task levels across different sessions and subjects, we proposed two domain adaptation methods, G-W and FG-W, for session-by-session and subject-by-subject alignment, respectively. The proposed methods can achieve the alignment accuracy greater than the chance level of 25$\%$. At the same time, the proposed methods do not require the same subset of fNIRS channels or further data interpolation for classification across all subjects and sessions as opposed to some other supervised methods like CNN and RNN. This will alleviate the pressure from having to exclude fNIRS channels that were noisy in one session but not in others, or from having to interpolate the signals to replace those noisy channels. Besides adapting domain adaptation method, we explored the effect of using the TARA signal processing algorithm for removing motion artifacts and found an improvement in the alignment accuracy results. In the future, we plan to explore the effect of our method on a larger sample of subjects and make it applicable for multiple source subjects.

\section{Appendix}

\begin{table}[H]
\centering
\caption{CNN architecture, where $d$ = number of channels (20 in our case), $w$ = number of time points (60 in our case), $T_1,T_2$ = length of time points after applying the filter and $C$ = number of classes (4 in our case).}
  \label{cnn_strucuture}
\begin{center}       
\begin{tabular}{|c|c|c|} 
\hline\hline
Layer & Operation  & Output Size \\
\hline\hline
Input  & --   &     (2, $d$, $w$)  \\
\hline
Conv2D  & 20 * filter (1, $10$)+BatchNorm+ReLU+Dropout(0.2) & ($20$, $d$, $T_1$) \\
\hline
Conv2D  & 20 * filter (1, $5$)+BatchNorm+ReLU+Dropout(0.2) & ($20$, $d$, $T_2$) \\
\hline
DepthwiseConv2D  & 20 * kernel ($d$, $1$)+BatchNorm+ReLU+Dropout(0.2) &($20$, $1$, $T_2$) \\
\hline
 --       &Flatten & ($20*T_2$) \\
\hline
 Dense $* 2$  &  --  &$C$\\
\hline\hline
\end{tabular}
\end{center}
\end{table} 

\begin{table}[h]
\caption{Number of retained channels for six subjects. The total number of channels is 20. "0" indicates when the particular session is removed.}
  \label{table:channel_number}
  \centering
  \begin{center}

  \begin{tabular}{|c|c|}
  \hline\hline
  Subject & Number of Retained Channels \\
  \hline
  Sub 1&[20, 20, 20, 0] \\
  \hline
  Sub 2 &[15, 17, 16, 16] \\
  \hline
  Sub 3&  [11, 0, 14, 8]  \\
  \hline
   Sub 4 & [20, 20, 20, 20]  \\
  \hline
   Sub 5 &[20, 20, 20, 20]\\
  \hline
  Sub 6 &  [0, 20, 20, 20] \\

   \hline\hline
  \end{tabular}
  \end{center}
\end{table}

\begin{table}[H]
\centering
\caption{Values of TARA parameters ($f_c$: cut-off frequency for the low-pass filter, $d$: order of the filter, $\theta$ and $\beta$:  regularization parameters for TARA, $\sigma$: noise standard deviation). Parameter values were chosen differently for $\Delta[HbO_2]$ and $\Delta[Hb]$ due to different noise level. For each subject, values of $\sigma$ and the choice of $\beta$ vary 
among sessions, as reported in square brackets "[]". "-" indicates when TARA is not applied or when the session is removed.}
\label{table:TARA_para}
\begin{center} 
\begin{tabular}{|c|c|c|c|c|c|c|}

\hline\hline
\multirow{2}{*}{Subject} & \multirow{2}{*}{Signal type} & \multicolumn{5}{c|}{Parameters} \\ \cline{3-7} 
            &                  & $f_c$ (Hz)    & $d$    & $\theta$    & $\beta$    & $\sigma$ $(\mu \text{M})$   \\ \hline\hline
\multirow{2}{*}{Sub 1}         &$\Delta[HbO_2]$& 0.15 &  1    &  0.01  &  [1.9, 1.9, 1.9, -]  &  [0.15, 0.15, 0.1, -]  \\ \cline{2-7} 
                               & $\Delta[Hb]$  & 0.15 &   1   &  0.01  &   [1.9, 1.9, 1.9, 1.9]  &   [0.05, 0.05, 0.05, 0.025]  \\ \hline
\multirow{2}{*}{Sub 2}         &$\Delta[HbO_2]$& 0.15  &  1    & 0.01  & [1.7, 1.6, 1.3, 1.2]  & [0.25, 0.23, 0.3, 0.15] \\ \cline{2-7} 
                               & $\Delta[Hb]$  & 0.15 &   1   & 0.01  &  [1.7, 1.7, 1.3, 1.3]    &  [0.06, 0.03, 0.04, 0.03]   \\ \hline
\multirow{2}{*}{Sub 3}         &$\Delta[HbO_2]$& 0.15 &   1   &  0.01    & [1.9, -, 1.5, 1.4]  & [0.15, -, 0.13, 0.15] \\ \cline{2-7} 
                               & $\Delta[Hb]$  & 0.15  &  1    &  0.01    & [1.9, -, 1.5, 1.6] &  [0.04, -, 0.05, 0.1]   \\ \hline
\multirow{2}{*}{Sub 4}         &$\Delta[HbO_2]$& -  &   -   & -    &   -   &  -   \\ \cline{2-7} 
                               & $\Delta[Hb]$  & -  &   -   &  -    &  -    &  -   \\ \hline
\multirow{2}{*}{Sub 5}         &$\Delta[HbO_2]$& 0.15 &   1   & 0.01  & [1.8, 1.3, 1.9, 1.9]  & [0.1, 0.1, 0.15, 0.14] \\ \cline{2-7} 
                               & $\Delta[Hb]$  & 0.15  &  1    & 0.01  & [1.8, 1.3, 1.9, 1.9]  &  [0.02, 0.015, 0.025, 0.02]   \\ \hline
\multirow{2}{*}{Sub 6}         &$\Delta[HbO_2]$& 0.15  &  1    &  0.01 & [-, 1.9, 1.9, 1.7]   & [-, 0.16, 0.1, 0.16] \\ \cline{2-7} 
                               & $\Delta[Hb]$  & 0.15  &  1    &   0.01     &   [-, 1.9, 1.6, 1.9]   &  [-, 0.03, 0.019, 0.015]   \\ \hline\hline
\end{tabular}
\end{center}
\end{table}




\subsection*{Disclosures}
The authors have no relevant financial interests in this manuscript and no potential conflicts of interest to disclose.

\subsection* {Acknowledgments}
This material is based upon work supported by the Air Force Office of Scientific Research under award number FA9550-18-1-0465. Any opinions, finding, and conclusions or recommendations expressed in this material are those of the author(s) and do not necessarily reflect the view of the United States Air Force. Shuchin Aeron would like to acknowledge support by NSF CAREER award CCF:1553075.

\subsection* {Code, Data, and Materials Availability}
No materials were used for the analysis. The code and data used to generate the results and figures are available in the Github repository:\par \url{https://github.com/boyanglyu/nback_align}.


\bibliography{report}   
\bibliographystyle{spiejour}   

\vspace{2ex}\noindent\textbf{Boyang Lyu} is a Ph.D. student at Tufts University under Prof. Shuchin Aeron. Her research involves unsupervised domain adaptation and signal processing. She has applied some of these techniques to word alignment and mental workload identification.

\vspace{2ex}\noindent\textbf{Thao Pham} is a Ph.D. student in the Diffuse Optical Imaging of Tissue (DOIT) Lab at Tufts University, under Prof. Sergio Fantini. Her research interests involve using near-infrared spectroscopy (NIRS) and coherent hemodynamics spectrocopy (CHS) model for noninvasive monitoring of cerebral blood flow (CBF) and cerebral hemodynamics in healthy human subjects and in clinical settings. 

\vspace{2ex}\noindent\textbf{Giles Blaney} is a Ph.D. student in the Diffuse Optical Imaging of Tissue (DOIT) lab at Tufts University under Prof. Sergio Fantini. He received a Bachelor of Science in Mechanical Engineering and Physics from Northeastern University in 2017, with minors in Electrical Engineering and Mathematics. Currently, Giles is researching methods for depth discrimination and imaging for use in near-infrared spectroscopy (NIRS). This includes studying the sensitivity of various NIRS optode arrangements in heterogeneous media, and development of the dual-slope method.

\vspace{2ex}\noindent\textbf{Angelo Sassaroli} received his Ph.D. degree in physics from the University of Electro-communications, Tokyo, Japan in 2002. From 2002 to 2007 he was a Research Associate at Tufts University, Medford, MA. Since 2007 he has been a Research Assistant Professor at Tufts University. He is co-author of more than 80 peer reviewed publications. His research interests focus on diffuse optical imaging.

\vspace{2ex}\noindent\textbf{Sergio Fantini} is Professor of Biomedical Engineering and principal investigator of the “Diffuse Optical Imaging of Tissue Laboratory” (DOIT Lab) at Tufts University. The DOIT Lab aims to develop non-invasive applications of near-infrared spectroscopy for medical diagnostics, monitoring of tissue oxygenation, quantitative assessment of tissue perfusion, and functional imaging. He co-authored with Dr. Irving Bigio a textbook on “Quantitative Biomedical Optics” published by Cambridge University Press. He is a Fellow of OSA, SPIE, and AIMBE. 

\vspace{2ex}\noindent\textbf{Shuchin Aeron} is an Associate Professor in the Dept. of ECE at Tufts University. Prior to Tufts he was a post-doctoral research scientist at Schlumberger Doll Research, Cambridge, MA form 2009-2011. He was awarded the School Of Engineering and Electrical and Computer Engineering Best Thesis Award. He is a recipient of the NSF CAREER award (2016). His main research interest lies at the intersection of information theory, statistical signal processing, optimization, and machine learning.

\vspace{1ex}
\noindent Biographies and photographs of the other authors are not available.

\listoffigures
\listoftables

\end{spacing}
\end{document}